\documentclass{article}
\usepackage[preprint]{neurips_2025}



\usepackage[utf8]{inputenc} 
\usepackage[T1]{fontenc}    
\usepackage{hyperref}       
\usepackage{url}            
\usepackage{booktabs}       
\usepackage{amsfonts}       
\usepackage{nicefrac}       
\usepackage{microtype}      
\usepackage{xcolor}         
\usepackage{amsmath, amssymb, amsfonts}
\usepackage{graphicx}
\usepackage{subcaption}
\usepackage{multirow}
\usepackage{float}
\usepackage{algorithm}
\usepackage{algpseudocode}

\usepackage{siunitx}  
\usepackage{caption}

\title{Differential Gated Self-Attention \\
\author{
Elpiniki Maria Lygizou \\
     TU Wien, Vienna, Austria \\
  \texttt{elpiniki.lygizou@tuwien.ac.at} \And
  Mónika Farsang\\
TU Wien, Vienna, Austria \\
\texttt{monika.farsang@tuwien.ac.at} 
\And
  Radu Grosu \\
TU Wien, Vienna, Austria \\
    \texttt{radu.grosu@tuwien.ac.at}} 
}

\begin{document}
\maketitle

\footnotetext{Code available at https://github.com/elyalyg/DGSA}

\begin{abstract}
Transformers excel across a large variety of tasks but remain susceptible to corrupted inputs, since standard self‐attention treats all query-key interactions uniformly. 
Inspired by lateral inhibition in biological neural circuits and building on the recent Differential Transformer’s use of two parallel softmax subtraction for noise cancellation, we propose \emph{Multihead Differential Gated Self-Attention (M-DGSA)} that learns per‐head input-dependent gating to dynamically suppress attention noise. Each head splits into excitatory and inhibitory branches whose dual softmax maps are fused by a sigmoid gate predicted from the token embedding, yielding a context-aware contrast enhancement. M-DGSA integrates seamlessly into existing Transformer stacks with minimal computational overhead. We evaluate on both vision and language benchmarks, demonstrating consistent robustness gains over vanilla Transformer, Vision Transformer, and Differential Transformer baselines. Our contributions are (i) a novel input-dependent gating mechanism for self‐attention grounded in lateral inhibition, (ii) a principled synthesis of biological contrast‐enhancement and self‐attention theory, and (iii) comprehensive experiments demonstrating noise resilience and cross-domain applicability.
\end{abstract}
\vspace{-2ex}
\section{Introduction}


%
%

Attention‐based Transformers \citep{vaswani2017attention} have revolutionized machine learning, driving breakthroughs in language, vision, and beyond. However, their uniform weighting of all query-key scores makes them vulnerable to input corruption - sensor noise in images or spurious tokens in text - which can be propagated and even amplified degrading performance in real‐world datasets. 
Existing techniques for enhancing attention robustness or reducing its complexity, typically either apply global regularization (e.g.\ dropout, weight decay) or rework the attention computation itself via sparsity or low-rank factorization \citep{wang2020linformer,choromanski2020rethinking,zaheer2020big}. More recently, the Differential Transformer (DIFF Transformer) \citep{ye2024differential} introduces a head-wise subtraction of two paired softmax maps to cancel common-mode noise, but its use of a single learned scalar limits its ability to adapt to low-level noise patterns that vary at the granularity of individual tokens.

In this work, we address the challenge of noise-cancellation in self‐attention by introducing \emph{Multihead Differential Gated Self‐Attention (M-DGSA)}, a lightweight module that learns an input-dependent inhibitory gating per‐head to dynamically adapt to and suppress attention noise. 

Inspired by \emph{lateral inhibition} in biological neural circuits, where neurons suppress neighboring activity to sharpen responses \citep{hartline1940receptive,kuffler1953discharge}, M-DGSA splits each head into excitatory and inhibitory branches. Dual softmax maps are fused via a sigmoid gate predicted from the token embedding, enabling fine‐grained contrast enhancement tailored to each query-key pair. M-DGSA integrates seamlessly into off‐the‐shelf Transformer-based stacks with negligible overhead. 

We evaluate M‐DGSA on both vision and language benchmarks - namely CIFAR‐10, CIFAR-100, FashionMNIST, SVHN, and ImageNet-1k for vision, Rotten Tomatoes, IMDB, AGNews, 20Newsgroups, and MNLI for language - under both clean and noisy settings. Empirically, M-DGSA consistently improves accuracy and noise resilience across these domains, outperforming vanilla Transformer \citep{vaswani2017attention}, Vision Transformer (ViT)~\citep{dosovitskiy2020image}, and Differential Transformer \citep{ye2024differential} baselines.
These results underscore the practical benefit of adaptive, input-dependent inhibition for attention‐based architectures for robust performance across diverse downstream tasks.

Our main contributions in this paper are as follows:
\begin{itemize}
    \item Introducing a per-head, input-conditioned gating mechanism that dynamically fuses the two parallel attention streams, enabling fine-grained, token-wise noise suppression,
    \item Adopting the lateral‐inhibition principles of biological systems' function into artificial attention mechanisms to provide an interpretable contrast-enhancement framework for self-attention,
    \item Validating M-DGSA across vision and language benchmarks, where it consistently boosts robustness and cross-domain adaptability under clean and corrupted conditions.
\end{itemize}




\section{Related Work}

\paragraph{Contrast-Enhancement and Inhibitory Motifs in Learning.}
In biological sensory neurons, lateral inhibition enhances spatial contrast by subtracting pooled neighboring activity, thereby sharpening feature edges and improving signal-to-noise ratio \citep{hartline1940receptive,kuffler1953discharge}. 

Inspirations from lateral inhibition have informed modern deep networks. These include Local Response Normalization (LRN) which applies fixed local competition to sharpen feature maps \citep{krizhevsky2012imagenet}; and spatial masking methods such as DropBlock, which attenuate contiguous activations to improve generalization \citep{ghiasi2018dropblock}. 
More recently, LI-CNN learns per-channel subtractive inhibition filters to dynamically emphasize salient patterns \citep{zhuang2023suppression}, Selective Kernel Networks (SKNet) employ channel-wise gating to adaptively fuse multiple convolutional kernels and dampen irrelevant features, mirroring lateral inhibition \citep{li2019selective}, 
while Gradient Mask applies inhibitory gating on gradients during backpropagation to filter gradient noise and stabilize training \citep{jiang2022gradient}.  
\paragraph{Attention Variants Towards Robustness.} Building on the Transformer \citep{vaswani2017attention} architecture, many attention variants have been proposed to improve efficiency and robustness. Low-rank methods such as Linformer \citep{wang2020linformer} and Performer \citep{choromanski2020rethinking} reduce complexity via kernel approximations, while sparsity-driven models like BigBird \citep{zaheer2020big} leverage block-sparse patterns. In vision, the Vision Transformer \citep{dosovitskiy2020image} demonstrates that pure self-attention over image patches, with an extra class token, can rival convolutional networks. ConViT \citep{d2021convit} then introduces locality-sensitive gating to blend convolutional inductive biases with self-attention, enabling dynamic control over spatial focus. Recently, the Differential Transformer \citep{ye2024differential} introduces head-wise subtraction of parallel softmax maps to cancel common-mode noise, improve spectral balance and reduce attention collapse. Multi-Token Attention~\citep{golovneva2025multi} performs a key–query convolution on the attention scores and applies a head-wise convolution across groups of attention heads to retrieve richer, reliable information from the input data.

Despite this rich history, to the best of our knowledge, head-wise, input-dependent lateral-inhibition remains unexplored in multi-head self-attention.

\section{Preliminaries}

\subsection{Scaled Dot-Product Attention}

In the original Transformer formulation \citep{vaswani2017attention}, given an input sequence of token embeddings  
\[
  X = [x_1, \dots, x_N] \in \mathbb{R}^{N \times d_{\mathrm{model}}},
\]
we compute query, key, and value matrices via learnable linear projections:
\[
  Q = X W_Q,\quad
  K = X W_K,\quad
  V = X W_V,
\]
where \(W_Q, W_K, W_V \in \mathbb{R}^{d_{\mathrm{model}}\times d}\).  
The standard scaled dot‐product attention then forms the weight matrix
\[
  A = \mathrm{softmax}\!\bigl(\tfrac{Q K^\top}{\sqrt{d}}\bigr)\;\in\mathbb{R}^{N\times N},
\]
applying the softmax row‐wise to normalize each query’s affinities, and the final output is
\[
  \mathrm{Attention}(Q,K,V) = A\,V \;\in\mathbb{R}^{N\times d}.
\]

\subsection{Differential Attention}

~\citep{ye2024differential} mitigate attention noise by contrasting two parallel attention maps.  For an input sequence 
\[
  X = [x_1, \dots, x_N]\;\in\;\mathbb{R}^{N \times d_{\mathrm{model}}},
\]

they first split queries and keys into paired streams and project values jointly:
\[
  [Q_1;Q_2] = X\,W_Q,\quad
  [K_1;K_2] = X\,W_K,\quad
  V = X\,W_V,
\]
with \(W_Q,W_K\in\mathbb{R}^{d_{\mathrm{model}}\times 2d'}\) and \(W_V\in\mathbb{R}^{d_{\mathrm{model}}\times 2d'}\).  Each stream then forms a standard scaled dot‐product attention,
\[
  A_i = \mathrm{softmax}\!\bigl(\tfrac{Q_i K_i^\top}{\sqrt{d}}\bigr)\;\in\mathbb{R}^{N\times N}, ,
  \quad i\in\{1,2\},
\]
and their outputs are combined via a learnable subtraction:
\[
  \mathrm{DiffAttn}(X) = (A_1 - \lambda\,A_2)\,V,\in\mathbb{R}^{N\times 2d'}.
\]
Here, \(\lambda\) is reparameterized to stabilize learning:
\[
  \lambda
  = \exp\bigl(\lambda_{q1} \lambda_{k1}\bigr)
  - \exp\bigl(\lambda_{q2} \lambda_{k2}\bigr)
  + \lambda_{\mathrm{init}},
\]
where \(\lambda_{q1},\lambda_{q2},\lambda_{k1},\lambda_{k2}\in\mathbb{R}^d\) are learnable vectors and \(\lambda_{\mathrm{init}} = 0.8 - 0.6\,e^{-0.3(l-1)}\) is a layer-dependent constant initialization for layer index \(l\in[1,L]\).

\subsection{Lateral Inhibition}

Lateral inhibition is a fundamental neural mechanism that enhances contrast by suppressing nearby activity. In rate‐based models, the neuron’s response is characterized by
\[
  r_i = \phi\Bigl(e_i - \alpha \sum_{j\in\mathcal N(i)} e_j\Bigr),
\]
where \(e_i\) is the direct excitatory input, \(\sum_j e_j\) aggregates neighboring signals, \(\alpha>0\) controls the inhibition strength, and \(\phi\) is a smooth nonlinearity.  This local subtractive process accentuates salient features while damping background noise, providing the theoretical underpinning for input-dependent gating in self-attention.
\vspace{-1ex}
\section{Method}

We propose \emph{Differential Gated Self‐Attention (M‐DGSA)}, the first self‐attention variant to embed an input-dependent lateral inhibition mechanism, combining the common-mode noise cancellation of differential attention with the adaptability of biological contrast‐enhancement.  
Building on the 
Differential Transformer \citep{ye2024differential} implementation (dual‐softmax streams, GroupNorm~\citep{wu2018group}, and $\lambda_{init}$ scheme), we replace its subtraction scalar with a lightweight gating network that learns input‐dependent and head‐specific inhibitory weights.

\subsection{Differential Gated Self‐Attention}


In our proposed \emph{Differential Gated Self-Attention}, the input tensor \(X\in\mathbb{R}^{N\times d_{model}}\) is linearly projected into two query-key substreams, \(Q^+,K^+\in\mathbb{R}^{N\times d'}\) (excitatory) and \(Q^-,K^-\in\mathbb{R}^{N\times d'}\) (inhibitory), alongside a unified value stream \(V\in\mathbb{R}^{N\times 2d'}\): 
\[
[\,Q^+;\,Q^-\,] = X\,W_Q,\quad
[\,K^+;\,K^-\,] = X\,W_K,\quad
V = X\,W_V,
\]
where \(W_Q,W_K,W_V\in\mathbb{R}^{d_{\mathrm{model}}\times 2d'}\) are the learnable weight matrices. 
Then, each substream computes its own scaled dot-product softmax attention map, denoted \(A^+\) and \(A^-\):
\[
  A^+ = \mathrm{softmax}\bigl(\tfrac{Q^+ {K^+}^\top}{\sqrt{d}}\bigr),
  \quad
  A^- = \mathrm{softmax}\bigl(\tfrac{Q^- {K^-}^\top}{\sqrt{d}}\bigr).
\]
Here we introduce a gating network, a per‐input (token) \(t\), per‐head \(h\) sigmoid gate 
\[
  g_{t,head_i} = \sigma\bigl(w_{g,head_i} x_t + b_{g,head_i}\bigr),
\]
which adaptively fuses excitation and inhibition streams
\begin{equation}
    A_{t,head_i} = g_{t,head_i}\,A^+_{t,head_i}\;-\;(1 - g_{t,head_i})\,A^-_{t,head_i}.
    \label{eq:attention}
\end{equation}

We emphasize subtracting the two gated-weighted terms, rather than blending them via weighted addition as in conventional gating mechanisms. By doing so, each position can dynamically choose how much of the inhibitory branch to apply, amplifying strong signals where needed and quelling noise elsewhere.

\begin{figure}[tb]
\noindent\begin{minipage}{.45\textwidth}
    \includegraphics[width=\linewidth]{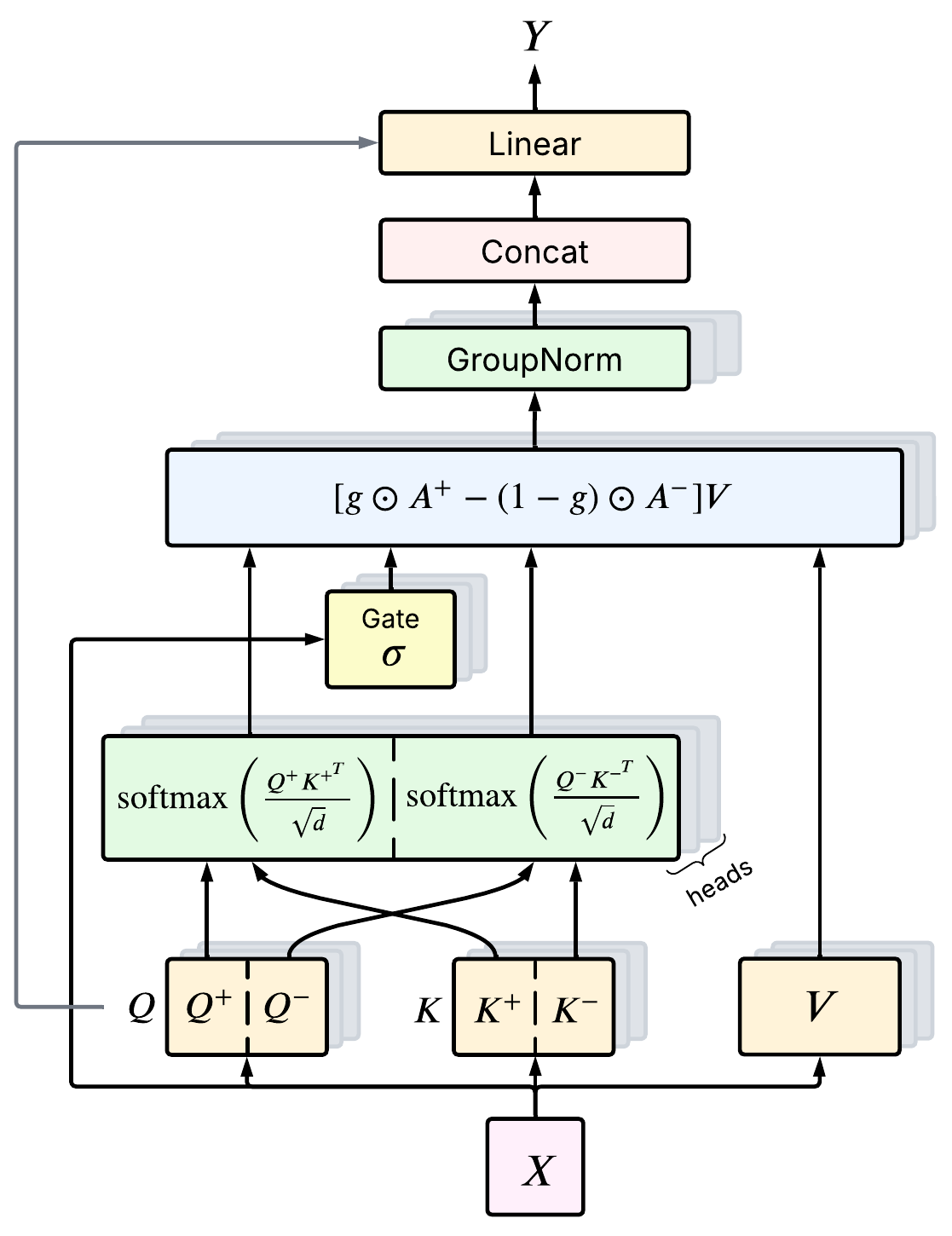}
    \label{fig:design}
\end{minipage}%
\hfill
\begin{minipage}[t!]{.5\textwidth}
  \centering
\label{alg:diff_attn}
\captionof{algorithm}{Multihead Differential Gated Self-Attention (M-DGSA)}
\begin{algorithmic}[1]
\Require $X\in\mathbb{R}^{N\times d_{model}}$, layer index $l$, weights $W_Q,W_K,W_V,W_g,W_o$, init $\lambda_{\mathrm{init}}$
\State $\lambda_l \gets 0.8 - 0.6\exp(-0.3(l-1))$
\State $Q \gets XW_Q,\quad K\gets XW_K,\quad V\gets XW_V$
\State Split $Q$ into $[Q^+;Q^-]$, split $K$ into $[K^+;K^-]$
\State $A^+ \gets \mathrm{softmax}(Q^+K^{+^{\top}}/{\sqrt d'})$
\State $A^- \gets \mathrm{softmax}(Q^-K^{-^{\top}}/\sqrt d')$
\State $g \gets \sigma(XW_g + b_g)$
\State $A \gets g\odot A^+ - (1-g)\odot A^-$
\State $H \gets (1-\lambda_l)\mathrm{GroupNorm}(AV)$
\State $O \gets$ reshape and concat heads of $H$
\State $Y \gets Q + OW_o$
\Ensure $Y\in\mathbb{R}^{N\times hd'}$
\end{algorithmic}
\end{minipage}
\vspace{-1ex}
\caption{Design of Multihead Differential Gated Self-Attention. The input tensor is projected into excitatory \((Q^+, K^+)\) and inhibitory \((Q^-, K^-)\) query-key pairs and shared values \(V\). Each branch generates its own softmax map \(A^+\) and \(A^-\), which are then "filtered" by a lightweight input-dependent headwise gating network that learns to attenuate noise - mimicking lateral inhibition in biological sensory systems. The fused attention is applied to \(V\) and each head’s output undergoes head-wise GroupNorm~\citep{wu2018group} before all heads are concatenated and linearly projected. An optional residual connection around the attention block can be added to improve gradient flow and training stability.}
\label{fig:design_alg}
\vspace{-2ex}
\end{figure}

\subsection{Multi-head Differential Gated Self‐Attention}

We extend the DGSA to \(h\) heads, as in the original Transformer multi-head attention mechanism \citep{vaswani2017attention}. Thereby, we tile the projection matrices into  
\[
[\mathbf{Q^+};\mathbf{Q^-}],\,[\mathbf{K^+};\mathbf{K^-}]\;\in\;\mathbb{R}^{N\times 2h\times d'}, 
\quad
\mathbf{V}\;\in\;\mathbb{R}^{N\times h\times 2d'},
\]
and then reshape into \(\{Q^\pm_{head_i},K^\pm_{head_i},V_{head_i}\}_{1}^h\), each of shape \(N\times 2\times d'\).  Each head independently computes its fused map \( A_{head_i}\) as in~\eqref{eq:attention} and used to attend over \(\mathbf{V}\), in the form of
\[
H_{head_i} = A_{head_i}\,V_{head_i}.
\]
This is followed by a GroupNorm~\citep{wu2018group} where RMSNorm \citep{zhang2019root} is applied in each head and scaled by \((1-\lambda_{init})\).


Concatenating \([head_1,\dots,head_i, \dots, head_h]\) along the embedding dimension and projecting with \(W_O\in\mathbb{R}^{hd'\times hd'}\) and adding an optional residual connection from $[\mathbf{Q^+};\mathbf{Q^-}]$ to preserve information flow in our model. This completes the design of our multi‐head attention block.


\subsection{DGT and DGViT: Transformer \& Vision Transformer Instantiations}
Our M-DGSA can be applied in any multi-head attention within  Transformer-based models. We explore two instantiations:
\vspace{-1ex}
\paragraph{Differential Gated Transformer (DGT).}
Starting from the standard Transformer encoder, we replace every multi-head self-attention block with M-DGSA. Each M-DGSA output is followed by a feed-forward network using the SwiGLU activation \citep{shazeer2020glu}.
Unlike in vision tasks, we found that an additional skip connection (shown in Fig.~\ref{fig:design_alg}) around the M-DGSA block did not improve convergence so we omit it in DGT.
\vspace{-1ex}
\paragraph{Differential Gated Vision Transformer (DGViT).}
Building on the ViT backbone, DGViT replaces each attention module with an M-DGSA layer, retaining the learnable class token in the attention computation and substitutes the GeLU activation \citep{hendrycks2016gaussian} in the feed-forward network with SwiGLU.
It also integrates an optional residual connection inside the M-DGSA block to facilitate gradient flow, stabilize training, and boost model's performance.

\vspace{-2ex}
\section{Experiments}
\vspace{-1ex}
We evaluate the proposed Multihead Differential Gated Self-Attention (M-DGSA) across two representative domains: vision and natural language classification. The complete experimental setup, hyperparameter schedules, and dataset details are provided in Appendix~\ref{appendix:exp_details}.

All models were trained from scratch (i.e., without pretrained weights) to isolate the contribution of the attention mechanism. As a result, absolute accuracies may be lower than prior work that fine-tunes pretrained backbones, however, this setting provides a controlled basis for cross-architecture comparison.
Experiments were implemented in PyTorch 2.x and executed on a single machine equipped with an NVIDIA A10G (24 GB) for smaller datasets, and on servers with NVIDIA A100 (40 GB and 80 GB) GPUs for ImageNet-1k.





\vspace{-1ex}
\subsection{Vision Classification}
\vspace{-1ex}

For vision classification, we leverage the built-in \texttt{torchvision.datasets} module, which provides standardized benchmarks including, CIFAR-10 and CIFAR-100 \citep{krizhevsky2009learning}, Fashion-MNIST \citep{xiao2017fashion}, and the Street View House Numbers (SVHN) dataset \citep{netzer2011reading}. For large-scale evaluation, we additionally include ImageNet-1k \citep{imagenet}. To assess robustness under challenging conditions, we further inject synthetic noise into the training images at multiple distortion levels. 

\vspace{-2ex}
\paragraph{Models.} 
For the four smaller benchmarks, comparisons include the original Vision Transformer (ViT) and our Differential Gated ViT (DGViT). We also adapt the Differential Transformer \citep{ye2024differential} to the vision setting, denoted DViT. In DGViT and DViT, the feed-forward module expands the hidden dimension by a factor of two or four, applies a SwiGLU activation, and then projects back to the base size. To study the role of regularization, we interleave one or two dropout layers between these projection and activation steps, thereby assessing the effect of different levels of stochastic masking during training.

For ImageNet-1k, we compare the performance of ViT, DViT, and DGViT under two configurations: a $4\times$ expansion with two dropout layers, which serves as our default, and a $16/3 \times$ expansion with two dropout layers, corresponding to the canonical setting of the Differential Transformer and providing a higher-capacity regime for comparison.

\vspace{-2ex}
\paragraph{Results on small benchmarks.} Table~\ref{tab:vis-results} reports accuracies averaged over five seeds on CIFAR-10, CIFAR-100, FashionMNIST, and SVHN. DGViT consistently outperforms the baseline ViT, with the largest improvement on CIFAR-10 (+2.2\%), highlighting the value of gated inhibition in low-resolution, noise-sensitive settings. On the other datasets, improvements are more modest (typically within 0.2-0.6\%), yet they remain steady across configurations and never come at the cost of degraded accuracy. When compared directly to DViT, DGViT yields competitive or superior performance in all settings. On CIFAR-100, where DViT underperforms relative to ViT, DGViT closes this gap and establishes a clear advantage. On FashionMNIST, DGViT slightly surpasses DViT, demonstrating that input-dependent gating remains beneficial even in simpler domains. On SVHN, DGViT matches or marginally improves over DViT, confirming that the gating mechanism does not erode the gains provided by differential attention.

These narrow but steady margins across diverse datasets suggest that the contribution of the gating mechanism is not limited to specific domains but rather yields a systematic robustness benefit, complementing differential attention without introducing instability. In practice, this reliability is as important as large improvements on single datasets, as it demonstrates that the method scales consistently across different visual conditions.
\vspace{-2ex}
\paragraph{Results on ImageNet-1k.} 
Table~\ref{tab:imagenet} summarizes performance at scale. Both DViT and DGViT substantially improve over the baseline ViT, with gains exceeding +6\% absolute accuracy. Between the two differential variants, DGViT achieves consistent, though moderate, improvements over DViT for both expansion factors. These findings indicate that input-dependent gating provides an additional layer of robustness on top of differential attention, yielding the strongest overall performance while preserving computational efficiency.
\vspace{-1ex}

\begin{table}[tb]
\caption{Classification accuracy (mean $\pm$ std) on small vision benchmarks. Averaged over 5 seeds.}
\label{tab:vis-results}
\begin{tabular}{lccccc}
  \toprule
  && CIFAR-10 (\%) & CIFAR-100 (\%)         & FashionMNIST (\%)      & SVHN (\%) \\
  \#Classes && 10 & 100 & 10 & 10 \\
  \#Size && 60,000 & 60,000 & 70,000 & 600,000 \\
  \midrule
  ViT  && $75.90\pm0.09$  & $47.01 \pm 0.56 $ & $93.15\pm0.11$   & $93.22\pm0.28$      \\
  \midrule
  \multicolumn{2}{l}{DViT} & & & \\
  \quad proj. & drop. & & \\
  \quad $\times 2$ & $\times 2$ & $76.76\pm0.28$	& $45.49\pm0.52$ & $93.39\pm0.10$ & $93.79\pm0.003$	 \\
  \quad $\times 4$ & $\times 1$ & $76.40\pm0.40$& $44.82\pm0.41$ & $93.32\pm0.001$ & $93.67\pm0.001$	 \\
  \quad $\times 4$ & $\times 2$ & $76.82 \pm 0.42$ & $45.55 \pm 0.01$ & $93.34\pm0.001$	 &	$93.80\pm0.003$ \\
  \quad $\times 16/3$ & $\times 2$ & $77.16\pm0.01$	& $45.66\pm0.26$ & $93.34\pm0.002$ &	 $93.99\pm0.001$\\
  \midrule
  \multicolumn{2}{l}{DGViT (ours)} & & & \\
  \quad proj. & drop. & & \\
  \quad $\times 2$ & $\times 2$  & $77.51\pm0.44$	& $47.13\pm0.26$ &	$93.42\pm0.07$	& $93.93\pm0.003$     \\
  \quad $\times 4$ & $\times 1$ & $77.77\pm0.29$	& $47.32\pm0.38$ &	$93.32\pm0.10$	& $\mathbf{94.16\pm0.004}$ \\
  \quad $\times 4$ & $\times 2$ & $78.06\pm0.41$	& $\mathbf{47.72\pm0.35}$ &	$\mathbf{93.45\pm0.13}$ &	$93.94\pm0.002$ \\
  \quad $\times 16/3$ & $\times 2$ & $\mathbf{78.31\pm0.01}$& $47.43\pm0.03$ &	$\mathbf{93.45\pm0.001}$ &$93.89\pm0.002$ \\

  \bottomrule
\end{tabular}
\end{table}

\begin{table}
    \caption{Classification accuracy (mean $\pm$ std) on ImageNet. Averaged over 5 seeds.}
    \centering
    \begin{tabular}{lcc}
    \toprule
    Model && Accuracy (\%) \\
    \midrule
         ViT && $ 59.94\pm0.04 $ \\
         \midrule
         \multicolumn{2}{l}{DViT} & \\
         \quad proj. & drop. &  \\
         \quad $\times 4$ & $\times 2$  & $ 65.75\pm0.01 $ \\
         \quad $\times 16/3$ & $\times 2$  & $66.66 \pm 0.006$\\
         \midrule
         \multicolumn{2}{l}{DGViT (ours)} & \\ 
         \quad proj. & drop. &  \\
         \quad $\times 4$ & $\times 2$  & $66.10\pm0.004$ \\
         \quad $\times 16/3$ & $\times 2$  & $\mathbf{67.08 \pm 0.003}$\\
  \bottomrule
    \end{tabular}
    \label{tab:imagenet}
\vspace{-3ex}
\end{table}

\subsection{Language Classification}
\vspace{-1ex}

For language classification, we design a progression of tasks that increase in difficulty with respect to the number of target classes and the diversity of linguistic structure. The evaluation begins with binary sentiment classification (Rotten Tomatoes~\citep{pang2005seeing}, IMDB~\citep{maas-EtAl:2011:ACL-HLT2011}), progresses to four-way news categorization task (AG News~\citep{del2005ranking}), and extends to the 20-class 20 Newsgroups benchmark ~\citep{twenty_newsgroups_113}, which demands fine-grained discrimination across diverse categories. To complement this controlled scaling, we additionally evaluate on MNLI~\citep{N18-1101}, a large-scale natural language inference benchmark spanning multiple genres, which serves as a rigorous test of generalization under heterogeneous conditions. Together, this suite of datasets provides a principled framework for assessing both the performance and robustness of our approach across varying levels of classification complexity.

\begin{table}[tb]
\captionof{table}{Classification accuracy (\%) (mean ± std) on language benchmarks. Averaged over 5 seeds.}
\label{tab:lang-results}
\resizebox{\textwidth}{!}{\begin{tabular}{lccccc}
  \toprule
  & Rotten Tom.   & IMDB         & MNLI & AGNews &  20 Newsg. \\
  \#Classes & 2 & 2 & 3 & 4  & 20\\
  \#Size & 10,000 & 50,000 & 433,000 & 127,600  & 18,000  \\
  \midrule
  Transformer &	$72.18\pm0.51$ & $85.34\pm0.22$ & $55.03\pm0.01$ &	$91.61\pm0.08$  &	$51.54\pm0.49$ \\
  \midrule
  DT & & & & &\\
  \quad proj. & & & & &\\
  \quad $\times 2$ & $73.99\pm0.91$ & $86.41\pm0.14$& $57.00\pm0.004$ & $91.85\pm0.23$ & $46.38\pm0.96$  \\
  \quad $\times 4$ & $73.06\pm0.70$ & $86.49\pm0.29$ & $56.11\pm0.01$ & $92.03\pm0.20$ & $47.38\pm2.08$\\
  \quad $\times 16/3$ & $73.19\pm0.85$ & $86.43\pm0.16$ &  $56.90\pm0.003$	& $92.00\pm0.19$ &		$49.99\pm2.80$\\  
  \midrule
  DGT (ours)  & & & &  &\\
  \quad proj. & & & &  &\\
 \quad $\times 2$ & $74.26\pm0.97$ & $86.47\pm0.17$ & $57.60\pm0.004$&	$\mathbf{92.25\pm0.14}$  &	$60.32\pm0.49$ \\
  \quad $\times 4$ &	$73.83\pm1.15$ & $\mathbf{86.97\pm0.04}$& $\mathbf{58.07\pm0.001}$ &	$91.89\pm0.14$  &	$59.07\pm0.48$  \\
  \quad $\times 16/3$  & $\mathbf{74.40\pm0.42}$  & $86.53\pm0.33$ & $57.26\pm0.004$&	$91.98\pm0.27$  &	$\mathbf{63.53\pm0.87}$  \\
  \bottomrule
\end{tabular}}
\vspace{-2ex}
\end{table}

\paragraph{Models.} We compare our DGT model’s performance to both the vanilla Transformer and the Differential Transformer (DT) under default configurations. To ensure fairness and to isolate the contribution of the attention mechanism, we adopt identical experimental conditions for DT and DGT: the feed-forward module expands the hidden dimension by factors of 2, 4, and 16/3 relative to the output dimension $Y$, applying SwiGLU before projecting back to the base dimension. This alignment guarantees that performance differences can be attributed to the gating mechanism rather than to architectural or training discrepancies.

\paragraph{Results.} Table~\ref{tab:lang-results} reports mean accuracy over five seeds. DGT improves over both baselines across all datasets. On the simpler sentiment tasks (Rotten Tomatoes, IMDB) and the four-class AG News, gains are modest (typically +1-2\%), reflecting that these benchmarks are near-saturated with strong baselines. Importantly, DGT maintains or improves performance across all expansion settings, indicating that input-dependent gating introduces no regressions even in low-complexity regimes. The advantages are most evident on 20 Newsgroups, where DGT consistently surpasses both baselines, demonstrating its capacity to handle fine-grained, high-variance classification with greater robustness. On MNLI, DGT consistently outperforms both baselines across expansion factors, with improvements of up to +3\% over the vanilla Transformer and +1-1.5\% over DT, underscoring its effectiveness in large-scale natural language inference.
Overall, the findings indicate that input-dependent gating yields consistent improvements on easier benchmarks and delivers pronounced advantages as task difficulty rises, underscoring its ability to scale reliably and adapt to diverse linguistic settings.

\subsection{Attention Visualization.}
We use the attention-rollout method \citep{abnar2020quantifying} to inspect how our models filter noise and sharpen structure on representative vision and language examples.

\begin{figure}[h!]
    \centering
    \begin{subfigure}[b]{0.75\textwidth}
        \includegraphics[width=\textwidth]{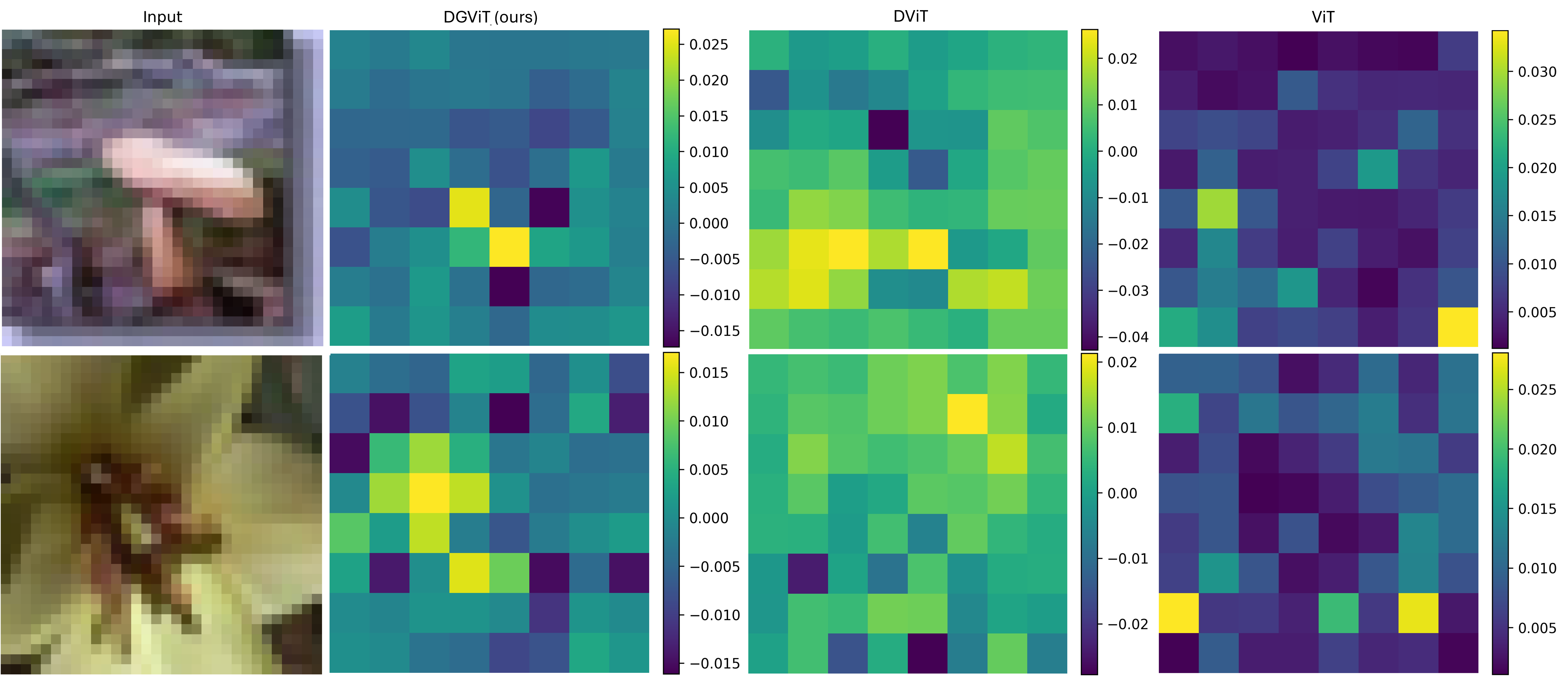}
        \label{fig:subfig1}
    \end{subfigure}
    \begin{subfigure}[b]{0.749\textwidth}
        \includegraphics[width=\textwidth]{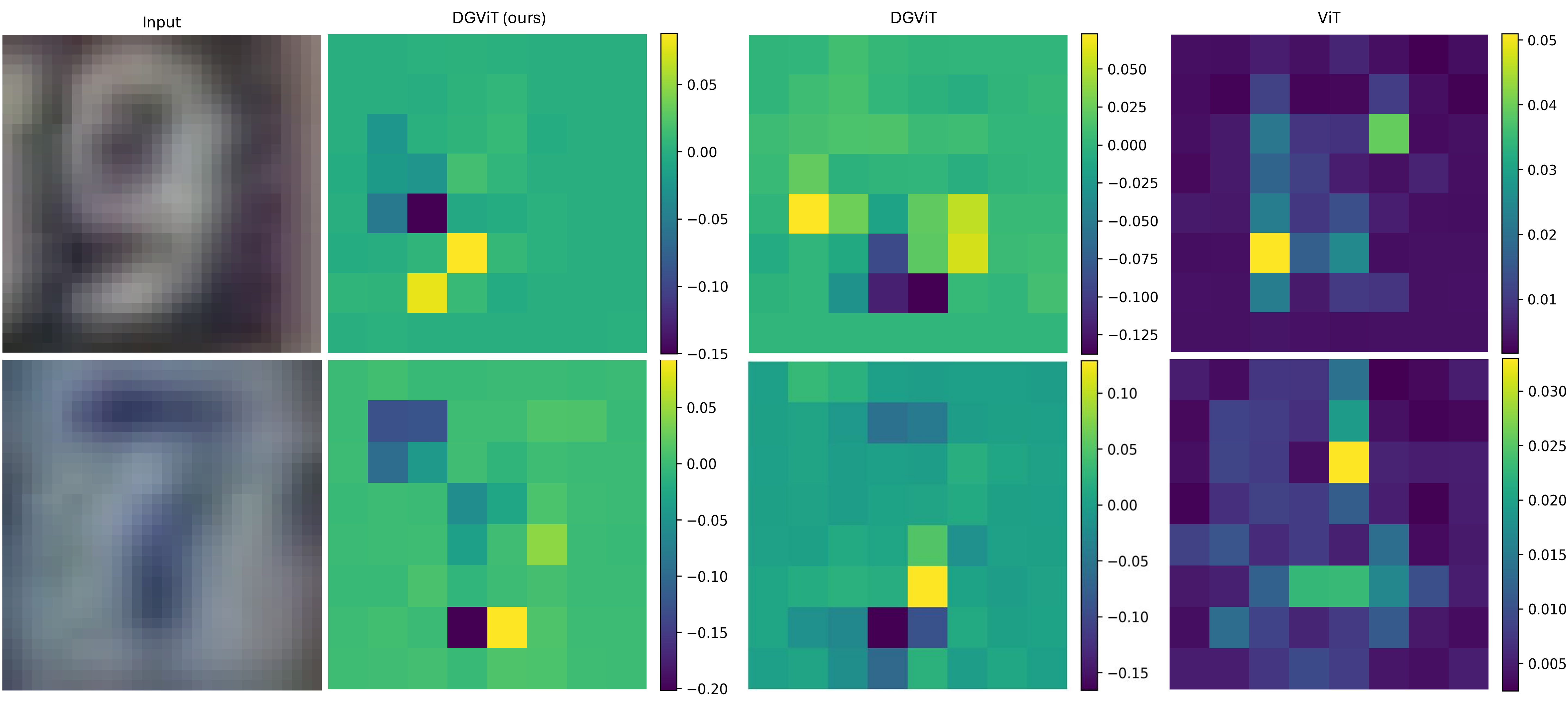}
        \label{fig:subfig3}
    \end{subfigure}
    \vspace{-2ex}
    \caption{Attention-rollout heatmaps for DGViT vs DViT vs ViT on CIFAR-100 (top two rows) and SVHN (bottom two rows). Bright yellow indicates high attention weights, whereas deep blue denotes low or suppressed attention. ViT disperses attention across both objects and background, DViT partially cancels common-mode noise but retains some distraction, while DGViT exhibits sharper boundary delineation and stronger alignment with task-relevant features.
    }
    \label{fig:three_figs}
\vspace{-3ex}
\end{figure}

\begin{figure}[h!]
  \centering
  \begin{subfigure}[b]{0.8\textwidth}
    \centering
    \includegraphics[width=\linewidth]{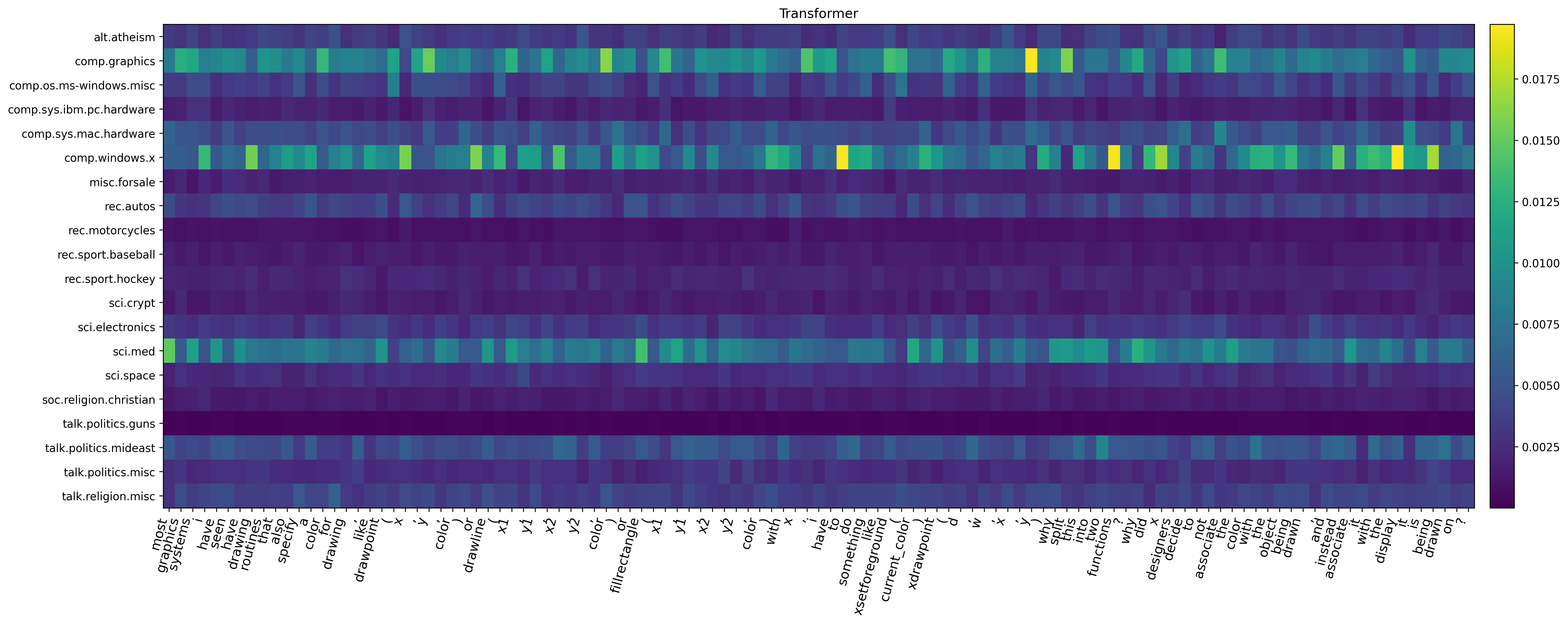}
    \caption{Transformer model attention}\label{fig:trans11}
  \end{subfigure}
  \begin{subfigure}[b]{0.8\textwidth}
    \centering
    \includegraphics[width=\linewidth]{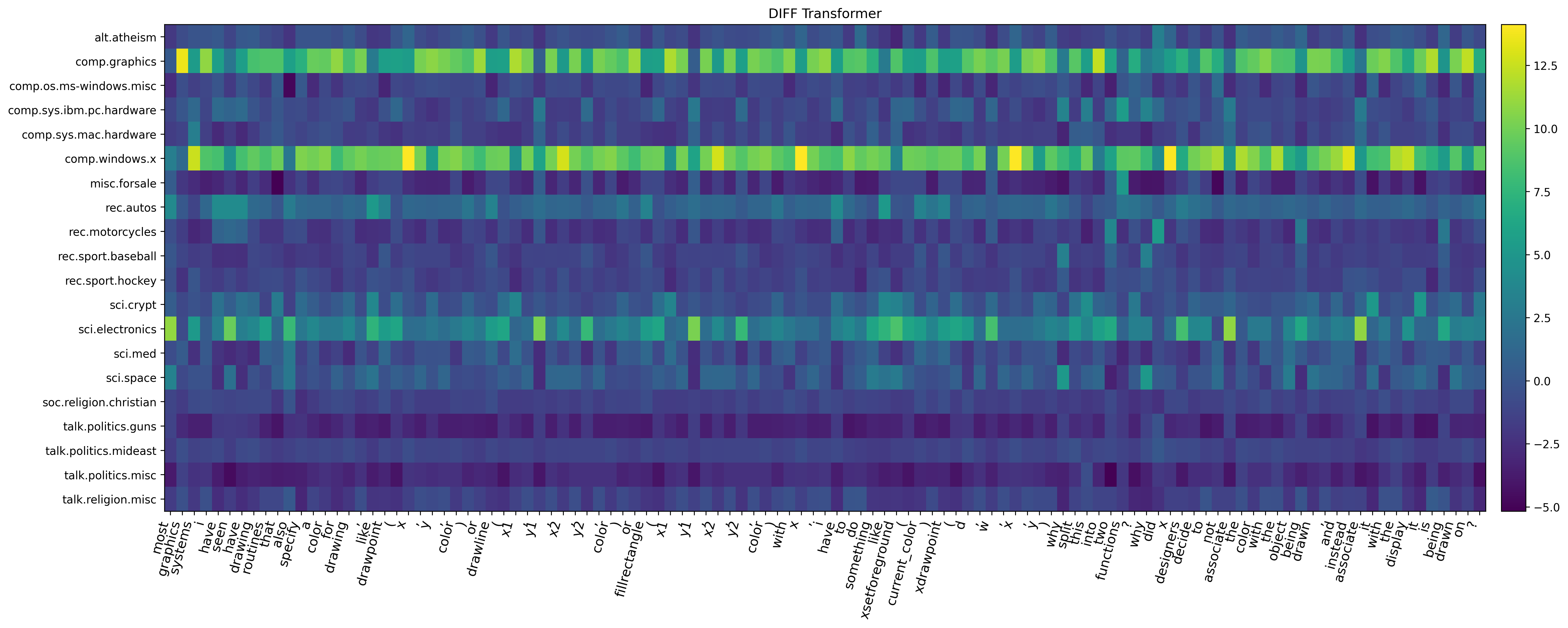}
    \caption{Differential Transformer model attention}\label{fig:dift11}
  \end{subfigure}
  \begin{subfigure}[b]{0.8\textwidth}
    \centering
    \includegraphics[width=\linewidth]{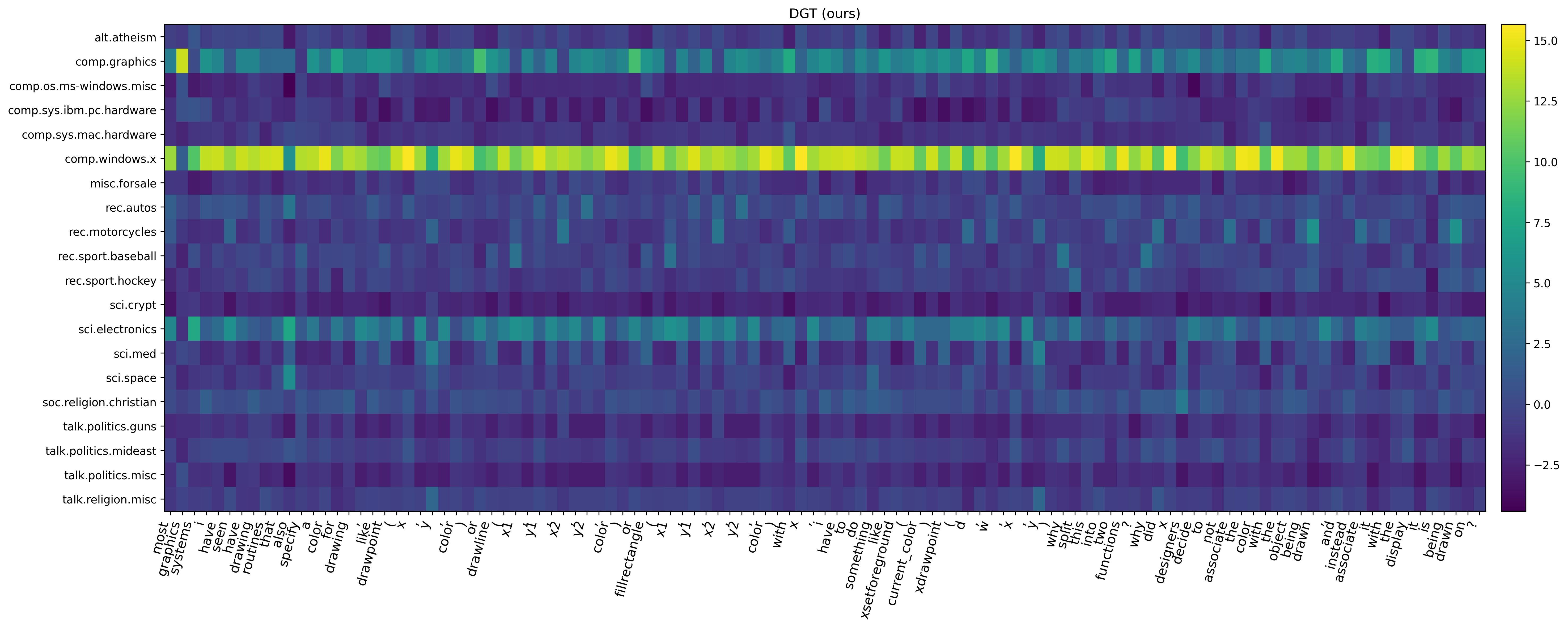}
    \caption{DGT (ours) model attention}\label{fig:dgt511}
  \end{subfigure}
  \caption{Attention-rollout for a 20 Newsgroups sample.  
Bright yellow indicates strong positive attention, and deep blue denotes inhibitory weights. 
(a) The standard Transformer baseline spreads its attention diffusely across both function words and content tokens, failing to isolate the class‐defining terms.
(b) The Differential Transformer partially suppresses common‐mode noise, but still assigns non‐negligible weight to other classes too.
(c) Our DGT model sharply concentrates on the key parts of the input while attenuating filler words. The focused, input-dependent gating aligns attention with the correct \textit{comp.windows.x} label and filters out background noise.}
  \label{fig:lang_attn}
\vspace{-3ex}
\end{figure}


\paragraph{Vision (DGViT vs Baselines).}  

Fig.~\ref{fig:three_figs} illustrates attention-rollout heatmaps for DGViT, DViT, and ViT on CIFAR-100 and SVHN. While ViT spreads attention broadly, often assigning high weights to background regions alongside the object of interest, DViT applies a differential contrast mechanism that partly sharpens focus but still leaves residual noise. 
DGViT, by contrast, produces the most discriminative patterns: excitatory weights are concentrated on semantically salient structures (e.g., mushroom contours or digit strokes), while inhibitory weights actively suppress background clutter, resulting in clearer boundaries and a more reliable focus on task-relevant regions.



\paragraph{Language (DGT vs Baselines).}
Fig.~\ref{fig:lang_attn} presents attention-rollout for (a) Transformer, (b) Differential Transformer, and (c) our DGT on a 20 Newsgroups sample, input text: \textit{"Most graphics systems I have seen have drawing routines that also specify a color for drawing, like Drawpoint(x,y,color) or Drawline(x1,y1,x2,y2,color) or Fillrectangle(x1,y1,x2,y2,color)…With X, I..."}, label: \textit{comp.windows.x}. The vanilla Transformer disperses its weights broadly over both common function words and various candidate labels, failing to single out the true class while the Differential Transformer partly suppresses common‐mode noise yet still spreads attention to competing classes, diluting its focus on the correct one. DGT’s input-dependent gating, in contrast,  concentrates focus on key class‐indicative words, filtering out irrelevant text and yielding a more interpretable, discriminative attention pattern, increasing attention mass on ground-truth and aligning sharply with \textit{comp.windows.x}.
\vspace{-1ex}


\section{Discussion}
\vspace{-2ex}
Our results demonstrate that embedding an input-conditioned lateral inhibition mechanism into self-attention yields substantial gains in noise resilience and generalization across both vision and language tasks. M-DGSA consistently sharpens attention maps, assigning strong excitatory weights to semantically relevant regions while suppressing background noise, without requiring any task-specific tuning. Its capacity to delineate fine boundaries suggests applications in safety‐critical domains - such as lane‐marking detection in autonomous driving or edge delineation in medical imaging - where robust, fine‐grained attention is paramount.
Despite these benefits, several limitations invite further exploration. We have so far evaluated primarily on mid- and large-scale benchmarks with synthetic noise, broadening the analysis to real-world distortions (e.g., motion blur, occlusions) may uncover additional strengths or weaknesses. Moreover, our study focused on self-attention, incorporating lateral inhibition into cross-attention - such as in vision-language models for VQA or image captioning - could enable dynamic, context-conditioned suppression of irrelevant inputs and further strengthen multimodal alignment. Looking forward, adapting lateral-inhibition gating to multimodal architectures, scaling up to natural corruption regimes, and developing hardware-efficient implementations that leverage the induced sparsity represent particularly promising directions.



\section{Conclusion}

We have introduced Multihead Differential Gated Self‐Attention (M‐DGSA), a lightweight yet powerful extension of multihead self‐attention that embeds per‐head input-dependent lateral inhibition into the dual‐softmax framework. Leveraging dynamic, context‐sensitive gating, M‐DGSA produces sharper and more selective attention maps, enhancing noise resilience and improving cross-domain performance on both vision and language benchmarks. Importantly, it integrates seamlessly into existing Transformer and ViT architectures with negligible overhead, making it directly applicable to a wide range of tasks.

Beyond empirical improvements, M‐DGSA advances the interpretability of attention mechanisms: its gating patterns mirror biological contrast‐enhancement, yielding clear visualizations of how models emphasize salient features and quench noise.  We view this biologically inspired paradigm as a significant step toward more robust, trustworthy attention models for next‐generation deep learning.


\section*{Reproducibility Statement}
We have taken several measures to facilitate reproducibility of our results. All datasets are standard public benchmarks with canonical train/test splits, requiring no additional preprocessing beyond what is provided in their official releases \ref{dsts}, \ref{appendix:data_splits}. Hyperparameter settings, model sizes, and training schedules for both vision and language tasks are detailed in Tables \ref{tab:hyper_vis}-\ref{tab:mnli_metrics} of the Appendix. Compute requirements and runtime estimates are reported in Appendix \ref{appendix:comprun}, and additional ablation studies on initialization and gate depth are provided in Appendix \ref{appendix:additional}. 
\section*{Acknowledgements}
This project has received funding from the European Union’s Horizon 2020 research and innovation programme under the Marie Skłodowska-Curie grant agreement No 101034277.

\bibliography{references}

\newpage
\appendix
\section*{Appendix}
\section{Additional Experimental Details}~\label{appendix:exp_details}
\vspace{-2em}
\subsection{Detailed description of the used datasets}
~\label{dsts}
\textbf{Rotten Tomatoes.}~\citep{pang2005seeing} $10,662$ movie reviews equally split between positive and negative sentiment.

\textbf{IMDB.}~\citep{maas-EtAl:2011:ACL-HLT2011} $50,000$ highly polarized reviews ($25,000$ train / $25,000$ test) for binary sentiment classification.

\textbf{MNLI.}~\citep{N18-1101} The Multi-Genre Natural Language Inference corpus is a dataset of 433,000 sentence pairs annotated with textual entailment, genre and label.

\textbf{AG News.}~\citep{del2005ranking} Derived from AG’s corpus of news articles, the AG News dataset comprises four categories - World, Sports, Business, and Science/Technology - by combining each article’s title and description and selecting the four largest classes. It contains $120,000$ training samples ($30,000$ per category) and $7,600$ test samples ($1,900$ per category).

\textbf{20 Newsgroups.}~\citep{twenty_newsgroups_113} $~18,000$ posts organized into 20 distinct newsgroups, each corresponding to a specific topic such as computer hardware, recreational activities, politics, science, or religion. We adopt the canonical split of $11,314$ training and $7,532$ test posts, yielding on average about $566$ training and $377$ test samples per class.

\textbf{CIFAR-10 / CIFAR-100.}~\citep{krizhevsky2009learning} $60,000$ colored images of size 32x32 pixels, evenly distributed across 10 distinct classes such as airplanes, automobiles, birds, cats, deers, dogs, frogs, horses, ships, and trucks. Each class includes $6,000$ images, with the dataset split into $50,000$ training and $10,000$ test images. CIFAR-100 is based on the same concept but with 100 classes and contains 600 samples for each class.

\textbf{FashionMNIST.}~\citep{xiao2017fashion} $70,000$ 28×28 pixel grayscale Zalando article images of 10 fashion item categories (e.g., t-shirt/top, trouser, bag, sandal). Maintaining the same structure and format as MNIST, FashionMNIST provides a standardized yet more complex alternative for evaluating machine learning models on image classification tasks.

\textbf{Street View House Numbers (SVHN).}~\citep{netzer2011reading} A large-scale, real-world image dataset comprising over 600,000 32×32-pixel digit crops extracted from Google Street View house numbers. It includes 10 classes (digits 0-9): 73,257 training images, 26,032 test images, and 531,131 additional “extra” samples of comparatively easier digits. By capturing digits in varied natural scenes - with diverse backgrounds, lighting conditions, and occlusions - SVHN poses a substantially more challenging recognition task than MNIST.

\textbf{ImageNet.}~\citep{imagenet} A large-scale, human-annotated image dataset organized according to the WordNet hierarchy, aiming to provide around 1000 images per concept, with tens of millions of labeled images overall. It was created to advance computer vision research by offering a high-quality benchmark for object categorization and supplying the large amounts of data needed for developing more generalizable machine learning methods.

\subsection*{Licenses for existing assets}

\begin{itemize}
    \item \textbf{Rotten Tomatoes}  
    ~\citep{pang2005seeing} Academic use only, see dataset terms at \url{https://nlp.stanford.edu/sentiment/index.html}
    \item \textbf{IMDB}  
    ~\citep{maas-EtAl:2011:ACL-HLT2011} Non-commercial academic use (see Stanford ACL IMDB dataset page).  
    \url{https://ai.stanford.edu/~amaas/data/sentiment/}
    \item \textbf{MLNI} OANC's license. \url{https://huggingface.co/datasets/nyu-mll/multi_nli}
    \item \textbf{AG News}  
    ~\citep{del2005ranking} Distributed under the CC BY-SA 3.0 license.  
    \url{https://www.di.unipi.it/~gulli/AG_corpus_of_news_articles.html}
    \item \textbf{20 Newsgroups}  
    ~\citep{twenty_newsgroups_113} Originally Usenet posts—distributed under the Creative Commons Attribution-ShareAlike 4.0 International (CC BY-SA 4.0).  
    \url{http://qwone.com/~jason/20Newsgroups/}
  \item \textbf{CIFAR-10 / CIFAR-100}  
    ~\citep{krizhevsky2009learning} Released under the MIT License.  
    \url{https://www.cs.toronto.edu/~kriz/cifar.html}
  \item \textbf{Fashion-MNIST}  
    ~\citep{xiao2017fashion} Public domain (MNIST) / MIT License (Fashion-MNIST).  
    \url{http://yann.lecun.com/exdb/mnist/},  
    \url{https://github.com/zalandoresearch/fashion-mnist}
  \item \textbf{SVHN}  
    ~\citep{netzer2011reading} Released under the MIT License.  
    \url{http://ufldl.stanford.edu/housenumbers/}
    \item \textbf{ImageNet}  
    ~\citep{imagenet} Non-commercial academic research and educational use.  
    \url{https://www.image-net.org/index.php}

\end{itemize}

\subsection{Dataset Splits}
~\label{appendix:data_splits}
For each language dataset, AG News, IMDB, 20 Newsgroups, and Rotten Tomatoes, we further split the training fold into $80\%$ train and $20\%$ validation (stratified by class). All final results are reported on the official test set.
For all vision benchmarks, CIFAR-10, CIFAR-100, Fashion-MNIST, and SVHN, we use the standard train/test partitions provided by each dataset and no further hold-out split was applied. 

\subsection{Hyperparameters}
~\label{appendix:hypera3}
\vspace{-2em}
\subsubsection{Vision Benchmarks}
For all vision datasets, we use the hyperparameters reported in Table~\ref{tab:hyper_vis}. For ImageNet-1k, we set the batch size to 64 and the patch size to 16.

\begin{table}[h!]
    \caption{Hyperparameters for vision datasets.}
    \centering
    \begin{tabular}{ll}
    \toprule
    Hyperparameter & Value \\
    \midrule
Epochs                               & 100                                                 \\
Batch size                           & 128                                                 \\
Learning rate                        & $3\times10^{-4}$                                    \\
Embedding dimension                  & 256                                                 \\
\# Layers                             & 8                                                   \\
\# Heads                              & 8                                                   \\
Dropout probability                  & 0.05                                                \\
Weight decay                         & 0.01                                                \\
Optimizer                            & AdamW ($\beta_{1}=0.9,\;\beta_{2}=0.999,\;\epsilon=10^{-8}$) \\
LR scheduler                         & CosineAnnealingLR                \\
Patch size                           & 4                                                   \\
Max sequence length                  & 100                                                 \\
\bottomrule
    \end{tabular}
    \label{tab:hyper_vis}
\end{table}




\subsubsection{Language Benchmarks}
For AG News, IMDB, and 20 Newsgroups, all text‐classification experiments use the hyperparameters listed in Table~\ref{tab:hyper_lang}. For Rotten Tomatoes, we instead employ 4 layers, a learning rate of $5 \times 10^{-4}$, and 500 warmup steps. The hyperparameters for MNLI are reported separately in Table~\ref{tab:hyper_mnli}.

\begin{table}[h!]
    \caption{Hyperparameters for small text datasets.}
    \centering
    \begin{tabular}{ll}
    \toprule
    Hyperparameter & Value \\
    \midrule

Epochs                               & 10                                                  \\
Batch size                           & 32                                                  \\
Learning rate                        & $3\times10^{-4}$                                    \\
Warmup steps                         & 1000                                                 \\
Max sequence length                  & 256                                                 \\
Minimum token frequency              & 2                                                   \\
Maximum vocabulary size              & 60,000                                             \\
Embedding dimension                  & 256                                                 \\
\# layers                            & 6                                                   \\
\# heads                             & 8                                                   \\
Dropout probability                  & 0.1                                                 \\
Optimizer                            & AdamW ($\beta_{1}=0.9,\;\beta_{2}=0.98,\;\epsilon\!=\!10^{-8}$) \\
Weight decay                         & 0.01 (0 for bias \& LayerNorm/RMSNorm weights)     \\
LR scheduler                         & Linear warmup (1000 steps) + linear decay           \\
Gradient clipping                    & Max-norm = 1.0                                      \\
        \bottomrule
    \end{tabular}
    \label{tab:hyper_lang}
\end{table}

\begin{table}[h!]
    \caption{Hyperparameters for MNLI dataset.}
    \centering
    \begin{tabular}{ll}
    \toprule
    Hyperparameter & Value \\
    \midrule

Epochs                               & 5                                                  \\
Batch size                           & 128                                                  \\
Learning rate                        & $3\times10^{-4}$                                    \\
Warmup steps                         & 4000                                                 \\
Max sequence length                  & 256                                                 \\
Minimum token frequency              & 2                                                   \\
Maximum vocabulary size              & 30,000                                             \\
Embedding dimension                  & 512                                                 \\
\# layers                            & 6                                                   \\
\# heads                             & 8                                                   \\
Dropout probability                  & 0.1                                                 \\
Optimizer                            & AdamW ($\beta_{1}=0.9,\;\beta_{2}=0.98,\;\epsilon\!=\!10^{-8}$) \\
Weight decay                         & 0.01 (0 for bias \& LayerNorm/RMSNorm weights)     \\
LR scheduler                         & Linear warmup (4000 steps) + linear decay           \\
Gradient clipping                    & Max-norm = 1.0                                      \\
        \bottomrule
    \end{tabular}
    \label{tab:hyper_mnli}
\end{table}

\subsection{Compute \& Runtime}
~\label{appendix:comprun}
ImageNet-1k experiments ran on NVIDIA A100
(40 GB and 80 GB) GPUs. All other experiments ran in PyTorch on a single NVIDIA A10G (24 GB).
\begin{itemize}
    \item Vision: ${\sim}0.5$min/epoch on CIFAR-10/100, ${\sim}0.6$ min on FashionMNIST, ${\sim}0.76$ min on SVHN, and ${\sim}31$min/epoch on ImageNet.
    \item Language: ${\sim}0.3$min/epoch on Rotten Tomatoes, ${\sim} 0.39$ min on IMDB, ${\sim}0.57$ min on AG News, ${\sim}0.5$ min on 20NewsGroups, and ${\sim}18$min/epoch on MNLI.
\end{itemize}

With DT and DGT, our goal was to avoid increasing - and in some cases even to reduce - the number of learnable parameters relative to the standard Transformer. This reduction stems from the use of SwiGLU, which splits projections in half, whereas the Transformer employs GeLU over the full hidden dimension.

The standard ViT has 5.9M parameters across CIFAR-10, CIFAR-100, FashionMNIST, and SVHN. For DViT and DGViT, the parameter count varies with the projection factor: a $\times2$ projection yields 3.8M parameters, while a $\times4$ projection increases this to 5.4M.

In the ImageNet experiments, all models maintain a comparable computational cost of 1.2 GFLOPs regardless of architecture. The baseline ViT has 5.7M learnable parameters and uses 2.7 GB of memory. DViT and DGViT with a $\times4$ projection keep the parameter count at 5.7M, but memory usage rises to 4.3 GB. The $\times16/3$ projection further increases parameters to 6.7M, while leaving the memory footprint unchanged. These results indicate that our architectural modifications primarily affect parameter count and memory requirements, while leaving overall FLOPs constant.



The number of trainable parameters in the language tasks of Rotten Tomatoes, IMDB, AGNews and 20 Newsgroups are reported in Table~\ref{tab:param_lang} and for the MNLI in Table~\ref{tab:mnli_metrics}.

\begin{table}[h!]
    \caption{Number of learnable parameters for language benchmarks.}
    \centering
    \begin{tabular}{lcccc}
    \toprule
  & Rotten Tomatoes   & IMDB     & AGNews &  20 Newsgroups \\
  \midrule
Transformer & $5.1$M & $27.6$M & $15.1$M & $13.3$M\\
\midrule
DT/DGT & & & &\\
 \quad  proj. $\times 2$ & $3.8$M & $25.7$M &	$13.2$M  &	$12$M  \\
  \quad  proj. $\times 4$ &	$4.6$M & $26.9$M &	$14.4$M  &	$12.8$M  \\
  \quad  proj. $\times 16/3$  & $5.1$M & $27.7$M &	$15.2$M  &	$13.4$M   \\
        \bottomrule
    \end{tabular}
    \label{tab:param_lang}
\end{table}

\begin{table}[H]
    \centering
    \caption{Metrics from the MNLI experiments.}
    \begin{tabular}{lccc}
    \toprule
         & No. of learnable params & Memory usage (in MB) & GFLOPs \\
         \midrule
         Transformer & 31.4M & 1.3k  & 4.4 \\
         \midrule
         DT/DGT &  &  & \\
         \quad  proj. $\times 2$ & 26.7M & 2.1k & 3.2 \\
         \quad proj. $\times 4$ & 31.4M & 2.1k & 4.4\\
         \quad  proj. $\times 16/3$ & 34.6M & 2.1k &  5.2\\
         \bottomrule
    \end{tabular}
    \label{tab:mnli_metrics}
\end{table}

\subsection{Additional Analyses}
~\label{appendix:additional}
\paragraph{Ablation on $\lambda_{init}$}
After experimenting with various $\lambda_{init}$ settings, we found that holding it fixed at 0.8 outperforms both other constant values and treating it as a learnable parameter.
\paragraph{Gate depth}
We observed that increasing the depth of the MLP gating layers consistently degraded the performance, therefore, our method employs a single-layer gating mechanism.



% ...

\clearpage
\bibliographystyle{abbrvnat}
\end{document}